%% file: main.tex
\documentclass[10pt,twocolumn,letterpaper]{article}

\usepackage[pagenumbers]{iccv} 


\definecolor{iccvblue}{rgb}{0.21,0.49,0.74}
\usepackage[pagebackref,breaklinks,colorlinks,allcolors=iccvblue]{hyperref}
\usepackage{multirow}
\usepackage{pifont}
\usepackage{makecell}
\usepackage{xcolor}

\definecolor{darkblue}{rgb}{0.0, 0.0, 0.5}

\def\paperID{8544} 
\def\confName{ICCV}
\def\confYear{2025}

\title{BANet: Bilateral Aggregation Network for Mobile Stereo Matching}

\author{Gangwei Xu$^{1}$, ~~Jiaxin Liu$^{1}$, ~~Xianqi Wang$^{1}$, ~~Junda Cheng$^{1}$, ~~Yong Deng$^{2}$\\
~~Jinliang Zang$^{2}$, ~~Yurui Chen$^{2}$, ~~Xin Yang$^{3,1}$\footnotemark[2]\\
[2mm]
{\normalsize $^1$Huazhong University of Science and Technology \quad $^2$Autel Robotics  \quad $^3$Optics Valley Laboratory
}}

\begin{document}
\maketitle
\input{sec/0_abstract}    

\renewcommand{\thefootnote}{\fnsymbol{footnote}}
\footnotetext[2]{Corresponding author.}

\input{sec/1_intro}

\input{sec/2_related}
\input{sec/3_method}
\input{sec/4_experiment}

\input{sec/5_conclusion}

{
    \small
    \bibliographystyle{ieeenat_fullname}
    \bibliography{main}
}

\input{sec/X_suppl}
\end{document}

%% file: sec/0_abstract.tex
\begin{abstract}
State-of-the-art stereo matching methods typically use costly 3D convolutions to aggregate a full cost volume, but their computational demands make mobile deployment challenging. Directly applying 2D convolutions for cost aggregation often results in edge blurring, detail loss, and mismatches in textureless regions. Some complex operations, like deformable convolutions and iterative warping, can partially alleviate this issue; however, they are not mobile-friendly, limiting their deployment on mobile devices. In this paper, we present a novel bilateral aggregation network (BANet) for mobile stereo matching that produces high-quality results with sharp edges and fine details using only 2D convolutions. Specifically, we first separate the full cost volume into detailed and smooth volumes using a spatial attention map, then perform detailed and smooth aggregations accordingly, ultimately fusing both to obtain the final disparity map. 
Experimental results demonstrate that our BANet-2D significantly outperforms other mobile-friendly methods, achieving 35.3\% higher accuracy on the KITTI 2015 leaderboard than MobileStereoNet-2D, with faster runtime on mobile devices. 
Code: \textcolor{magenta}{https://github.com/gangweix/BANet}.
\end{abstract}

%% file: sec/1_intro.tex
\section{Introduction}
\label{sec:intro}

\begin{figure}[t]
\centering
{\includegraphics[width=1.0\linewidth]{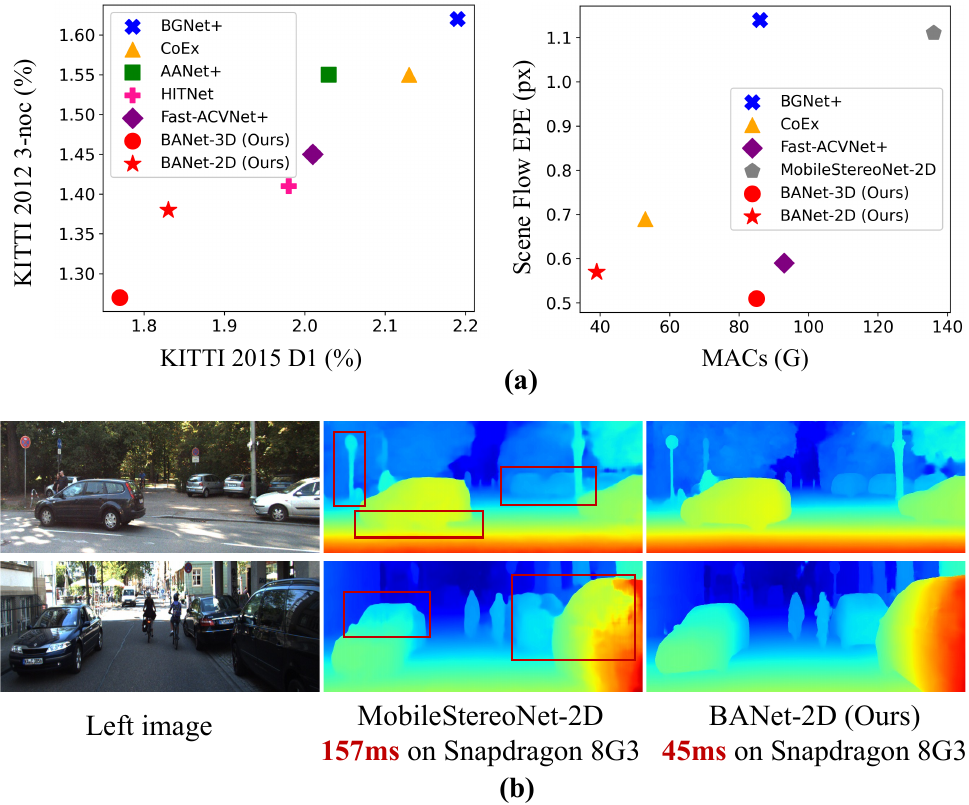}}
\caption{\textbf{(a)}: Comparison with top-performing real-time methods (high-end GPUs)~\cite{hitnet,aanet,fast-acv,coex,bgnet}. 
\textbf{(b)}: Visual results and latency on Qualcomm Snapdragon 8 Gen 3 (8G3). MobileStereoNet-2D~\cite{mobilestereonet} blurs edges, loses details, and causes mismatches in textureless regions due to pure 2D convolutions for cost aggregation. In comparison, our bilateral aggregation effectively addresses these issues while remaining mobile-friendly, eliminating the need for complex operations~\cite{aanet,hitnet}. The latency is measured for input stereo pairs with a resolution of $512 \times 512$.}
\label{fig:teaser}
\vspace{-10pt}
\end{figure}

Metric depth estimation plays a critical role in a wide range of real-world applications, such as drone navigation, smartphone photography, and robotic surgery. It can be broadly categorized into tasks include stereo matching~\cite{igev++}, depth completion~\cite{promptdepth,zhu2025svdc,xiang2025depthor}, and monocular depth estimation~\cite{hu2024metric3d}, and so on. Among them, stereo matching focuses on finding pixel correspondences between left and right images, enabling depth recovery via triangulation.
Currently, deep learning-based methods~\cite{monster,croco-stereo,unistereo,zerostereo,openstereo,guo2024stereo,foundationstereo,jiang2025defom} have dominated stereo matching or depth benchmarks, consistently setting new state-of-the-art results on public leaderboards~\cite{kitti2015,Middlebury,eth3d}. Despite significant advancements, deep stereo matching methods still face challenges when deployed on mobile and embedded devices. 
These challenges include: 1) computational and memory demands exceeding mobile platforms' capabilities; 2) difficulties in deploying complex or customized operators~\cite{aanet,hitnet}; and 3) the trade-off where simpler models tend to sacrifice accuracy. In this paper, we aim to explore a mobile-friendly stereo matching method that achieves both real-time inference speed and high accuracy using only mobile-friendly operations.

Recently, to improve stereo matching performance, some works~\cite{gwcnet,acvnet,pcwnet} have designed sophisticated cost volume representations, which play a crucial role in the stereo matching pipeline. For example, GwcNet~\cite{gwcnet} designs a group-wise correlation volume that computes correlations group by group and then concatenates them. ACVNet~\cite{acvnet,fast-acv} introduces an attention concatenation volume that transforms the correlation volume into attention weights, which are then used to filter the concatenation volume~\cite{gcnet,psmnet}. PCWNet~\cite{pcwnet} proposes a pyramid combination and warping cost volume. These cost volume representations, in conjunction with a large number of 3D convolutions, provide a notable improvement in terms of prediction accuracy. However, the accompanying high computational and memory costs make it nearly impossible to deploy on mobile devices. 

To accelerate the runtime of stereo matching, subsequent works have adopted downsampled or sparse cost volume representations~\cite{stereonet,bgnet,fast-acv,decomposition,deeppruner}, as well as lightweight aggregation networks~\cite{iinet, coex, hdrflow, cgi}. For example, StereoNet~\cite{stereonet} and BGNet~\cite{bgnet} construct a low-resolution cost volume (1/8 of the image resolution), while DeepPruner~\cite{deeppruner} and Fast-ACVNet~\cite{fast-acv} create sparse cost volumes by pruning the candidate disparity range. These methods can achieve real-time performance on high-end GPUs; however, they still require stacked 3D convolutions to regularize the cost volumes, which are not mobile-friendly and make it difficult for these methods to meet real-time demands on mobile devices, such as those powered by Qualcomm Snapdragon chips.

An alternative solution is to replace 3D convolutions with 2D convolutions for cost volume aggregation; however, This can lead to edge blurring at disparity discontinuities or a loss of fine details. To mitigate these issues, AANet~\cite{aanet} adopts deformable convolutions~\cite{deformable} to perform adaptive cost aggregation. HITNet~\cite{hitnet} introduces iterative warping operations to progressively restore the edges and fine details of the disparity map. Unfortunately, these complex operations, such as deformable convolutions and iterative warping, are not mobile-friendly and are quite expensive to deploy on mobile devices. Contrary to these methods, MobileStereoNet-2D~\cite{mobilestereonet} utilizes pure 2D convolutions for cost aggregation; however, it exhibits a severe degradation in accuracy, as shown in \cref{fig:teaser}. To this end, a motivating question arises: \emph{How to design a mobile-friendly stereo network while maintaining high prediction accuracy}?

In this paper, we propose a novel bilateral aggregation network (BANet) that simultaneously achieves real-time performance on mobile devices and high prediction accuracy. Considering that images contain both high-frequency detailed regions and low-frequency smooth/textureless regions, which are essential for achieving accurate predictions, directly applying standard 2D convolutions for cost aggregation across all regions tends to blur edges, lose details, and produce mismatches in textureless areas. To address this, the proposed BANet first separates the full cost volume into a detailed and smooth cost volume, then aggregates each individually, and finally fuses them. Through this bilateral aggregation, comprising a detailed aggregation branch and a smooth aggregation branch, our method achieves state-of-the-art performance while preserving clear edges and fine details in complex scenes, as shown in \cref{fig:teaser}. 

Furthermore, accurately identifying high-frequency detail regions and low-frequency smooth regions is also crucial for final prediction accuracy. For this purpose, we propose a new scale-aware spatial attention module to differentiate between high-frequency details and edges, and smooth regions. The features of different scales have varying perceptions and receptive fields. Fine-scale features can perceive more high-frequency detail information, while coarse-scale features capture more low-frequency smooth information. Our scale-aware spatial attention takes full advantage of features at different scales to produce an accurate spatial attention map that effectively separates detailed and smooth regions.

To demonstrate the effectiveness of our approach, we conduct extensive experiments on the Scene Flow~\cite{dispNetC} and KITTI~\cite{kitti2012,kitti2015} datasets. Our pure 2D convolution-based BANet-2D outperforms other lightweight methods~\cite{coex,bgnet,fast-acv,hitnet} on the KITTI 2012 and 2015 leaderboards. We examine the latency on Qualcomm Snapdragon 8 Gen 3. As shown in \cref{fig:teaser}, our BANet-2D is mobile-friendly and takes only 45ms for input stereo pairs with a resolution of $512 \times 512$, delivering high-quality results that preserve edges and details, significantly outperforming MobileStereoNet-2D~\cite{mobilestereonet}. We also extend BANet to a 3D version, and the resulting BANet-3D model achieves the highest prediction accuracy among all real-time methods (on high-end GPUs) on the KITTI 2012 and 2015 leaderboards. 

In summary, our main contributions are as follows:
\begin{itemize}
    \item We present a novel bilateral aggregation network for mobile stereo matching that achieves high-quality results using only 2D convolutions.
    
    \item We propose a scale-aware spatial attention module that accurately identifies high-frequency details and edges, and low-frequency smooth regions.
    \item Our approach can run in real-time on mobile devices with high prediction accuracy, significantly outperforming other mobile-friendly methods. 
    \item The extended 3D version, BANet-3D, achieves the highest accuracy among all real-time methods on GPUs.
\end{itemize}

%% file: sec/2_related.tex
\section{Related Work}
\label{sec:related}

\subsection{Deep Stereo Methods}
Recently, deep stereo methods can primarily be categorized into two types: cost volume aggregation-based approaches~\cite{psmnet,gwcnet,acvnet,ganet,pcwnet,zhang2024exploring,acfnet,lsp,cheng2024adaptive,stereorisk} and iterative optimization-based approaches~\cite{raft-stereo,igev,anystereo,dlnr,mc,crestereo,zhao2024hybrid,xu2023memory, jing2023uncertainty,mocha,los,selective,eglcr,monster,jing2025stereo,bartolomei2025stereo}. A representative method within the first category is PSMNet~\cite{psmnet}. PSMNet constructs a 4D concatenation volume and employs a stacked hourglass network, composed of 3D convolutions, to aggregate this volume. Due to the simplicity and excellent performance of PSMNet, many subsequent works have attempted to improve it in terms of cost volume construction and cost aggregation. For example, GwcNet~\cite{gwcnet} proposes group-wise correlation volume, ACVNet~\cite{acvnet} introduces attention concatenation volume, and PCWNet~\cite{pcwnet} presents pyramid combination and warping volume. These stereo matching methods enhance the representational capacity of the cost volume, leading to improved accuracy. However, they typically come with an expensive computational cost. To improve efficiency, Cascade-Stereo~\cite{cascade} and CFNet~\cite{cfnet} propose cascade cost volume representations, constructing the cost volume in a coarse-to-fine manner. 

For cost aggregation, GA-Net~\cite{ganet} introduces two guided aggregation layers to replace the widely used 3D convolutional layer, while CoAtRS~\cite{coatrsnet} proposes global attention along the disparity dimension for more comprehensive aggregation. However, these methods incur high computational costs, making real-time deployment challenging even on high-end GPUs, let alone on mobile devices.

Iterative optimization-based methods~\cite{raft-stereo,crestereo} iteratively update disparity using matching features retrieved from a correlation volume, thus avoiding the computationally expensive cost aggregation operations. However, they typically require a large number of iterations to obtain an optimal disparity. To improve optimization efficiency and accuracy, IGEV~\cite{igev,igev++} introduces a more comprehensive geometry encoding volume, from which matching features are iteratively indexed to update the disparity. Despite its excellent performance, it still struggles to be deployed on mobile devices due to the iterative indexing operations.

\subsection{Real-time Stereo Methods}
To speed up stereo matching inference time, many methods~\cite{stereonet,bgnet,deeppruner,fadnet,chang2020attention} directly construct and aggregate a deeply downsampled cost volume, such as 1/8 of the image resolution. However, these downsampled cost volumes can lead to a significant degradation in accuracy. To maintain comparable accuracy, CoEx~\cite{coex} still constructs a high-resolution cost volume but uses a more lightweight aggregation network. Fast-ACVNet~\cite{fast-acv}, on the other hand, introduces a high-resolution sparse attention module that only computes sparse matches at a high resolution. However, these methods achieve real-time inference only on high-end GPUs. The extensive use of 3D convolutions makes them challenging to deploy on mobile devices.

To replace costly 3D convolutions while maintaining comparable accuracy, AANet~\cite{aanet} leverages deformable 2D convolutions to enable adaptive cost aggregation, thereby alleviating the well-known edge-fattening issue. Unlike aggregation-based approaches, HITNet~\cite{hitnet} avoids constructing an explicit cost volume and instead progressively recovers a full-resolution disparity through iterative warping operations. However, complex operations such as deformable convolutions and iterative warping are generally not mobile-friendly, making them challenging to deploy on mobile devices.

MobileStereoNet-2D~\cite{mobilestereonet} employs 2D MobileNet blocks~\cite{mobilenetv2} for cost aggregation, which are mobile-friendly. However, images typically contain information at varying frequencies, and objects exhibit different disparities. As a result, 2D aggregation often leads to edge blurring, detail loss, and mismatches in textureless regions. In contrast, our bilateral aggregation adaptively separates the cost volume based on the corresponding frequency information or disparities, and then performs targeted aggregation for each.

%% file: sec/3_method.tex
\section{Bilateral Aggregation Network}
\label{sec:method}
\begin{figure*}
    \centering
    \includegraphics[width=1.0\linewidth]{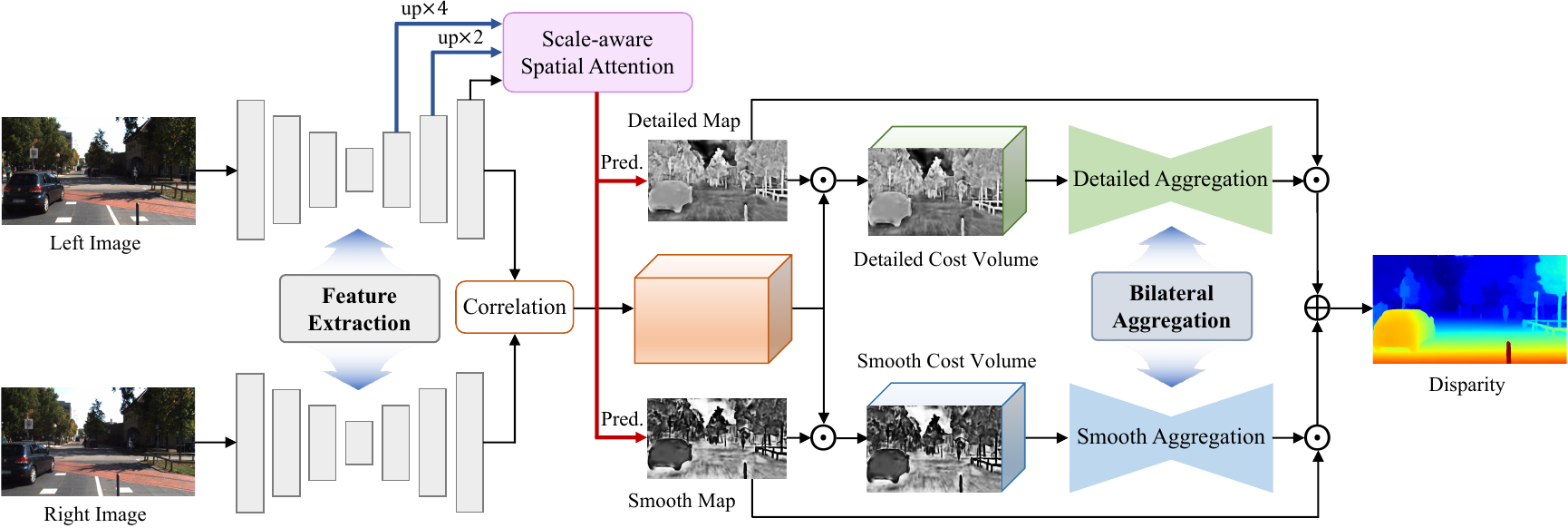}
    \caption{Overview of our proposed Bilateral Aggregation Network (BANet).To effectively handle both high-frequency detailed regions and low-frequency smooth regions, we use detailed and smooth maps to separate the full cost volume into detailed and smooth volumes. This enables targeted aggregation for each, with the detailed and smooth volumes processed independently. The detailed map highlights high-frequency detailed regions, while the smooth map highlights the opposite. We also introduce a new scale-aware spatial attention to more accurately identify detailed and smooth regions within the image.}
    \label{fig:network}
\end{figure*}

In this section, we introduce the detailed structure of the proposed BANet, illustrated in \cref{fig:network}. It consists of four steps: feature extraction, correlation volume construction, bilateral aggregation, and disparity prediction.
Most existing methods use 3D convolutions to aggregate cost volumes, which improves accuracy but is computationally expensive and unsuitable for mobile devices. In contrast, 2D convolutions are lightweight but often cause blurring and mismatches. Our proposed bilateral aggregation achieves high accuracy using only efficient, mobile-friendly operations.

\subsection{Bilateral Aggregation}
\label{sec:ba}
An image typically contains both high-frequency detailed regions and low-frequency smooth or textureless regions. 
Therefore, using only a 2D aggregation network to aggregate the entire cost volume makes it difficult to manage both detailed and smooth regions, often resulting in edge blurring, detail loss, and mismatches in textureless regions (as shown in \cref{fig:teaser}). To address these issues, we propose bilateral aggregation, which first separates the full cost volume into detailed and smooth volumes. It then uses a detailed aggregation branch for the detailed volume and a smooth aggregation branch for the smooth volume (as shown in \cref{fig:network}).

Specifically, given a full correlation volume $\mathbf{C}_{cor}$ constructed through simple feature correlation, we separate it into a detailed cost volume $\mathbf{C}_{d}$ and a smooth cost volume $\mathbf{C}_{s}$. This separation operation is based on a spatial attention map $\mathbf{A}$ introduced in \cref{sec:ssa},
\begin{equation}
\begin{aligned}
\mathbf{C}_{d} = & \; \mathbf{A} \odot \mathbf{C}_{cor},\\
\mathbf{C}_{s} = & \; (1-\mathbf{A}) \odot \mathbf{C}_{cor},
\end{aligned}
\end{equation}
where $\odot$ represents the Hadamard Product. The spatial attention map $\mathbf{A}$ highlights the high-frequency detailed regions, while the map $(1-\mathbf{A})$ highlights the low-frequency smooth regions.

After separating into detailed and smooth cost volumes, we accordingly employ a detailed aggregation branch $\mathbf{G}_{d}$ to aggregate the detailed cost volume and a smooth aggregation branch $\mathbf{G}_{s}$ for the smooth cost volume,
\begin{equation}
\begin{aligned}
\mathbf{C}_{d}^{'} = & \; \mathbf{G}_{d} (\mathbf{C}_{d}),\\
\mathbf{C}_{s}^{'} = & \; \mathbf{G}_{s} (\mathbf{C}_{s}).
\end{aligned}
\end{equation}

For simplicity, we adopt the same structure for both $\mathbf{G}_{d}$ and $\mathbf{G}_{s}$, but do not share their weights. Here, we only detail $\mathbf{G}_{d}$. Following previous work, the detailed aggregation network $\mathbf{G}_{d}$ consists of a series of inverted residual blocks~\cite{mobilenetv2}: 4 blocks at 1/4 resolution, 6 blocks at 1/8 resolution, and 8 blocks at 1/16 resolution. Each inverted residual block contains point-wise, depth-wise, and another point-wise 2D convolution, with an expansion factor of 4. The channel numbers for the cost features at 1/4, 1/8, and 1/16 resolutions are 32, 64, and 128, respectively.

The detail aggregation network targets high-frequency regions, while the smooth aggregation network focuses on low-frequency, textureless regions for disparity predictions. Finally, we fuse the aggregated detailed cost volume and the smooth cost volume by:

\begin{equation}
\begin{aligned}
\mathbf{C}_{agg} = & \; \mathbf{A} \odot \mathbf{C}_{d}^{'} + (1-\mathbf{A}) \odot \mathbf{C}_{s}^{'}.
\end{aligned}
\end{equation}

\textbf{Extension to 3D.} Naturally, we can extend the concept of bilateral aggregation to 3D convolutions, further boosting the accuracy of our model. The 3D aggregation network includes three down-sampling blocks and three up-sampling blocks. Each down-sampling block contains two $3 \times 3 \times 3$ kernel-sized 3D convolutions, while each up-sampling block includes a $4 \times 4 \times 4$ kernel-sized 3D transposed convolution followed by two $3 \times 3 \times 3$ kernel-sized 3D convolutions. Our 3D extension, BANet-3D, achieves the highest accuracy among all published real-time methods on high-end GPUs.

\subsection{Scale-aware Spatial Attention}
\label{sec:ssa}
Fine-scale image features capture more high-frequency details and edges, while coarse-scale features encompass more low-frequency, smooth, and textureless information. 
Therefore, to accurately separate detailed regions and smooth regions, we propose a scale-aware spatial attention module that learns the differences in multi-scale image features to generate an attention map. This map effectively distinguishes detailed regions and smooth regions.

As shown in \cref{fig:network}, the multi-scale left image features $\mathbf{F}_{l,16}$, $\mathbf{F}_{l,8}$, and $\mathbf{F}_{l,4}$ are scaled to 1/4 resolution before being input into our spatial attention module. The scaled features are represented as $\mathbf{F}_{l,16}^{up}$, $\mathbf{F}_{l,8}^{up}$, and $\mathbf{F}_{l,4}$. 
First, we apply a convolutional layer to each scaled feature to obtain intermediate features with the same number of channels, and then we concatenate them. Second, we use another convolutional layer followed by a $sigmoid$ activation function to predict the spatial attention map. In this way, the attention map $\mathbf{A}$ is obtained by:

\begin{equation}
\begin{aligned}
\mathbf{S} = & \; Concat([Conv(\mathbf{F}_{l,16}^{up}),Conv(\mathbf{F}_{l,8}^{up}),Conv(\mathbf{F}_{l,4})]),\\
\mathbf{A} = & \; \sigma(Conv(\mathbf{S})),
\end{aligned}
\end{equation}
where $Concat$ indicates the concatenation operator, $Conv$ represents the 2D convolutional operator, and $\sigma$ denotes the $sigmoid$ function.

As shown in \cref{fig:network} and \cref{fig:ba}, the spatial attention map $\mathbf{A}$ effectively highlights high-frequency details and edges, as these regions typically exhibit high feature values and distinct variations within the scale-aware perception. By applying a reverse operation, we obtain an inverse spatial attention map $(1-\mathbf{A})$ that highlights low-frequency smooth and textureless regions.

\subsection{Network Architecture}

\textbf{Feature Extraction.} Given the left image $\mathbf{I}_{l}\in\mathbb{R}^{{3}\times{H}\times{W}}$ and the right image $\mathbf{I}_{r}\in\mathbb{R}^{{3}\times{H}\times{W}}$, we employ a pre-trained MobileNetV2 on ImageNet \cite{imagenet} as our backbone, extracting multi-scale feature maps at 1/4, 1/8, 1/16, and 1/32 of the original resolution, respectively. Starting from the 1/32 resolution image features, we iteratively apply up-sampling blocks until reaching a 1/4 resolution.
In more detail, each up-sampling block employs a transpose convolution with a $4 \times 4$ kernel and a stride of 2, followed by a $3 \times 3$ kernel-sized convolution. Finally, we obtain multi-scale left image features: $\mathbf{F}_{l,4}$ at 1/4 resolution, $\mathbf{F}_{l,8}$ at 1/8 resolution, and $\mathbf{F}_{l,16}$ at 1/16 resolution, which are then used for scale-aware spatial attention generation, while $\mathbf{F}_{l,4}$ and $\mathbf{F}_{r,4}$ are used for correlation volume construction, as shown in \cref{fig:network}.

\textbf{Correlation Volume Construction.} Given the left feature map $\mathbf{F}_{l,4}$ and right feature map $\mathbf{F}_{r,4}$, we construct the correlation volume by,
\begin{equation}
\mathbf{C}_{cor}(d,x,y)=\frac{1}{N_c}\langle\mathbf{F}_{l,4}(x,y), \mathbf{F}_{r,4}(x-d,y)\rangle, 
\end{equation}
where $\langle \cdot, \cdot\rangle$ is the inner product of two feature vectors, $N_c$ denotes the number of channels, and $d$ represents the disparity level.

\textbf{Disparity Prediction.} We use the proposed bilateral aggregation to aggregate the correlation/cost volume, which is detailed in \cref{sec:ba}. After obtaining the aggregated cost volume, we apply the softmax operation to it to regress the disparity map $\mathbf{d}_0$:
\begin{equation}
\mathbf{d}_{0} = \sum\limits_{d=0}^{D_{max}/4-1} d \times Softmax(\mathbf{C}_{agg}(d)),
\end{equation}
where $D_{max}$ denotes the predefined maximum disparity value, and $d$ represents the predefined disparity range from 0 to $D_{max}/4-1$. The disparity map $\mathbf{d}_0$ has a size of $H/4 \times W/4$. We use superpixel weights~\cite{superpixel} around each pixel in the left image for a weighted combination of local neighboring points in $\mathbf{d}_0$, resulting in a full-resolution disparity map $\mathbf{d}_1 \in \mathbb{R}^{{H}\times{W}}$.

\subsection{Loss Function}
The entire network is trained in a supervised, end-to-end manner, with the final loss function defined as follows:
\begin{equation}
    \mathcal{L} = \lambda_{0}Smooth_{L_1}(\mathbf{d}_0-\mathbf{d}_{gt})+\lambda_{1}Smooth_{L_1}(\mathbf{d}_1-\mathbf{d}_{gt})
\end{equation}
where $\mathbf{d}_{gt}$ denotes the ground-truth disparity map, and $Smooth_{L_1}$ represents Smooth L1 loss.

%% file: sec/4_experiment.tex
\section{Experiments}
\label{sec:experiment}

\begin{table*}
  \centering
  \begin{tabular}{l|l|cc|ccc}
    \toprule
    Agg. Type & Model & \makecell{Bilateral \\ Aggregation} &\makecell{{Scale-aware} \\ {Spatial Attn}} & EPE (px) & Bad 3.0 (\%) & MACs (G) \\ 
\midrule
\multirow{3}{*}{2D Conv} & Baseline & \ding{55} & \ding{55}  & 0.63 & 2.75 & 29 \\
& BA & \ding{51} & \ding{55} & 0.59 & 2.57 & 38 \\
& BA+SSA (BANet-2D) & \ding{51} & \ding{51} & \textbf{0.57} & \textbf{2.49} & 39 \\
\midrule
\multirow{3}{*}{3D Conv} & Baseline & \ding{55} & \ding{55}  & 0.56 & 2.43 & 57 \\
& BA & \ding{51} & \ding{55} & 0.53 & 2.27 & 80 \\
& BA+SSA (BANet-3D) & \ding{51} & \ding{51} & \textbf{0.51} & \textbf{2.21} & 85 \\

\bottomrule
\end{tabular}
  \caption{Ablation study on the Scene Flow test set. We integrate bilateral aggregation into two types of aggregation networks: 2D aggregation networks and 3D aggregation networks. Results demonstrate that our bilateral aggregation can significantly improve the accuracy of existing aggregation networks.}
  \label{tab:ablation}
  \vspace{-5pt}
\end{table*}

\begin{table} 
\setlength{\tabcolsep}{4.pt} %
    \centering
    \begin{tabular}{l|cc|ccc}
     \toprule
     \multirow{2}{*}{Method} & \multicolumn{2}{c|}{KITTI 2012} &\multicolumn{3}{c}{KITTI 2015}\\
     & 3-noc & 3-all & D1-bg & D1-fg & D1-all \\
     \midrule
     w/o BA & 1.77 & 2.21 & 1.85 & 3.67 & 2.15 \\
     BANet-2D &\textbf{1.38} &\textbf{1.79}  &\textbf{1.59} &\textbf{3.03} (\textbf{\textcolor{darkblue}{17\%$\uparrow$}}) & \textbf{1.83} \\
     \midrule
     w/o BA & 1.54 & 2.01  & 1.66 & 3.87 & 2.03 \\
     BANet-3D &\textbf{1.27} &\textbf{1.72} &\textbf{1.52} &\textbf{3.02} (\textbf{\textcolor{darkblue}{22\%$\uparrow$}}) &\textbf{1.77} \\
    \bottomrule
    \end{tabular}
    \caption{Ablation study on the test sets of KITTI 2012 and 2015. 
Our bilateral aggregation (BA) effectively enhances the prediction accuracy for both background (D1-bg) and foreground (D1-fg) regions, particularly in the foreground, which typically contains more high-frequency details and edges.}
\label{tab:ablation_kitti}
\end{table}

\begin{table}
  \centering
  \begin{tabular}{l|cc}
    \toprule
    Method & EPE (px) & MACs (G) \\ 
\midrule
DeepPruner-Fast~\cite{deeppruner} & 0.97 & 219 \\
StereoNet~\cite{stereonet} &1.10 & 104\\
BGNet+~\cite{bgnet} & 1.14 & 86 \\
MobileStereoNet-2D~\cite{mobilestereonet} & 1.11 & 136 \\
MobileStereoNet-3D~\cite{mobilestereonet} & 0.80 & 615 \\
CoEx~\cite{coex} & 0.67 & 53 \\
LightStereo-L~\cite{lightstereo} & 0.59 & 92\\
Fast-ACVNet~\cite{fast-acv} & 0.64 & 79\\
Fast-ACVNet+~\cite{fast-acv} & 0.59 & 93 \\
\midrule
BANet-2D (Ours) & 0.57 & \textbf{39} \\
BANet-3D (Ours) & \textbf{0.51} & 85 \\
\bottomrule
\end{tabular}
  \caption{
  Quantitative evaluation on the Scene Flow test set. 
  The MACs are measured for an input size of $960\times540$.}
  \label{tab:sceneflow}
\vspace{-10pt}
\end{table}

\begin{table*}
    \centering
    \begin{tabular}{l|cccccc|ccc|c}
    \toprule
     \multirow{3}{*}{Method} & \multicolumn{6}{c|}{KITTI 2012 \cite{kitti2012}} & \multicolumn{3}{c|}{ KITTI 2015 \cite{kitti2015}}  & \multirow{3}{*}{MACs (G)}  \\
    & 3-noc & 3-all & 4-noc & 4-all & \thead{EPE \\ noc} & \thead{EPE\\all} & D1-bg & D1-fg & D1-all &  \\
    \midrule
    {DispNetC~\cite{dispNetC}} & 4.11 & 4.65 & 2.77 & 3.20 & 0.9 & 1.0 & 4.32 & 4.41 & 4.34 & - \\
     {AANet+~\cite{aanet}} & 1.55 & 2.04  & 1.20 &1.58 & 0.4 & 0.5 &  1.65 & 3.96 & 2.03 & - \\
    {DecNet~\cite{decomposition}}  & - & - & - & - & - & - & 2.07 & 3.87 & 2.37 & - \\
    {BGNet+~\cite{bgnet}} & 1.62 & 2.03 &  1.16 & 1.48 & 0.5 & 0.6 & 1.81 & 4.09  & 2.19 & 76 \\
    {CoEx~\cite{coex}} & 1.55 & 1.93 &  1.15 & 1.42 & 0.5 & 0.5 & 1.79 & 3.82  & 2.13 & 49 \\
    {DeepPruner-Fast\cite{deeppruner}} & - & - & - & - & - & - & 2.32 & 3.91 & 2.59 & 194 \\
    {HITNet~\cite{hitnet}}  & 1.41 & 1.89 &  1.14 & 1.53 & 0.4 & 0.5 & 1.74 & 3.20  & 1.98 & \underline{47} \\
    {Fast-ACVNet+~\cite{fast-acv}} & 1.45 & 1.85 &  1.06 & 1.36 & 0.5 & 0.5 & 1.70 & 3.53  & 2.01 & 85 \\
    {Fast-ACVNet~\cite{fast-acv}} & 1.68 & 2.13 &  1.23 & 1.56 & 0.5 & 0.6 & 1.82 & 3.93  & 2.17 & 72 \\
    MobileStereoNet-2D~\cite{mobilestereonet} & - & - & - & - & - & - & 2.49 & 4.53 & 2.83 & 127 \\
    MobileStereoNet-3D~\cite{mobilestereonet} & - & - & - & - & - & - & 2.75 & 3.87 & 2.10 & 564 \\
    \midrule
    BANet-2D (Ours) & \underline{1.38} & \underline{1.79} & \underline{1.01} & \underline{1.32} & 0.5 & 0.5 & \underline{1.59} & \underline{3.03} & \underline{1.83} & \textbf{36} \\
    BANet-3D (Ours) & \textbf{1.27} & \textbf{1.72} & \textbf{0.95} & \textbf{1.27} & 0.5 & 0.5 & \textbf{1.52} & \textbf{3.02} & \textbf{1.77} & 78 \\
    \bottomrule
    \end{tabular}
    \caption{Quantitative evaluation on the test sets of KITTI 2012~\cite{kitti2012} and KITTI 2015~\cite{kitti2015}. Previous methods reported their runtime on their own GPUs; however, the runtime can vary across different GPU models. For a fair comparison, we measure MACs, which are consistent across GPU models. The MACs are measured for an input size of $1242\times375$. \textbf{Bold}: Best, \underline{Underline}: Second best.}
    \vspace{-3pt}
    \label{tab:kitti_benchmark}
\end{table*}


\begin{figure}
\centering
{\includegraphics[width=1.0\linewidth]{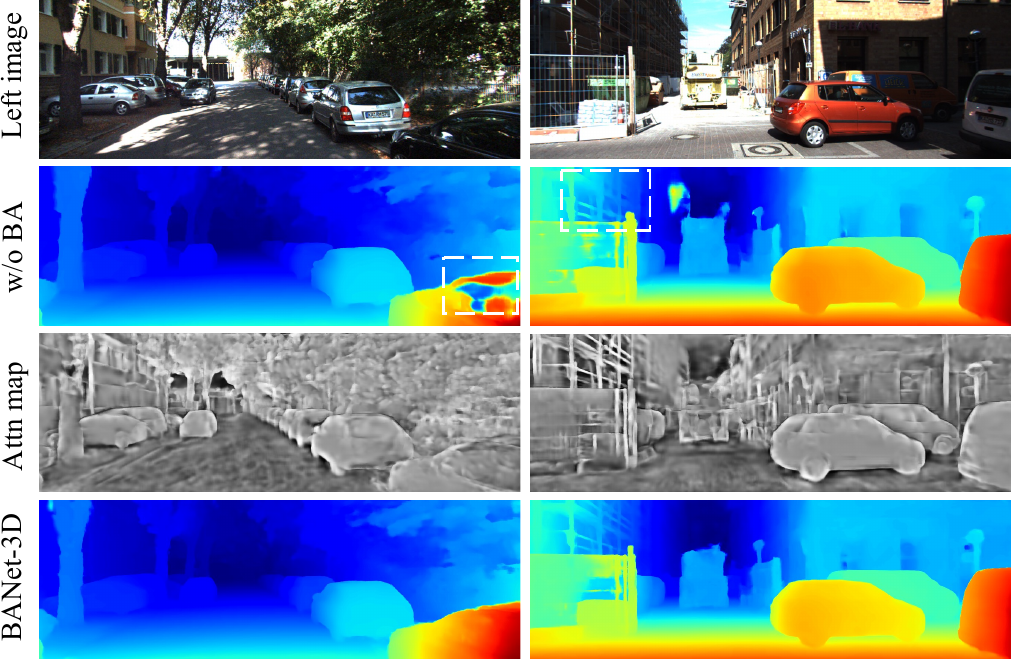}}
\caption{Our bilateral aggregation (BA) significantly enhances performance in both detailed and textureless regions. The Attention map highlights the high-frequency detailed regions. 
Significant improvements are highlighted by white dashed boxes.}
\label{fig:ba}
\vspace{-10pt}
\end{figure}

\subsection{Datasets and Evaluation Metrics}
\textbf{Scene Flow}~\cite{dispNetC} is a synthetic dataset, consisting of Flyingthings3D, Driving, and Monkaa. The dataset provides 35,454 training pairs and 4,370 testing pairs of size $960\times540$ with dense disparity maps. The Scene Flow dataset provides two versions: Cleanpass and Finalpass. We use the Finalpass of Scene Flow for training and testing since it is more like real-world images than the Cleanpass, which contains more motion blur and defocus. The end-point error (EPE) and disparity outlier rate Bad 3.0 are used as the evaluation metrics. Bad 3.0 is defined as the percentage of pixels with disparity error (EPE) greater than 3 pixels.

\textbf{KITTI 2012}~\cite{kitti2012} and \textbf{KITTI 2015}~\cite{kitti2015} are datasets for real-world driving scenes. KITTI 2012 contains 194 training pairs and 195 testing pairs, and KITTI 2015 contains 200 training pairs and 200 testing pairs. Both datasets provide sparse ground-truth disparities obtained with LiDAR. We submit our predicted disparity maps to the KITTI website to obtain quantitative evaluation results. For KITTI 2012, we report the percentage of pixels with errors larger than x disparities in both non-occluded (x-noc) and all regions (x-all), as well as the overall EPE in both non-occluded (EPE-noc) and all the pixels (EPE-all). For KITTI 2015, we report the percentage of pixels with EPE larger than max(3px, 0.05$\mathbf{d}_{gt}$) in background regions (D1-bg), foreground regions (D1-fg), and all (D1-all).

\subsection{Implemention Details}
We implement our approaches with PyTorch and perform our experiments using NVIDIA RTX 3090 GPUs. We first train our approaches on the Scene Flow dataset for 200k steps with a batch size of 16, and then fine-tune the pre-trained Scene Flow model on a mixed dataset of KITTI 2012 and 2015 training sets for 50k steps. During training, images are randomly cropped to a size of $256\times512$. For all experiments, we use the AdamW~\cite{adamw} optimizer with a one-cycle learning rate schedule, where the maximum learning rate is set to 8e-4. In our experiments, we set $\lambda_{0}$ and $\lambda_{1}$ to 0.3 and 1.0, respectively. $D_{max}$ is set to 192.

\subsection{Ablation Study}
We conduct extensive ablation studies on the Scene Flow~\cite{dispNetC} and KITTI~\cite{kitti2012,kitti2015} datasets to validate the effectiveness of the proposed approaches. The proposed bilateral aggregation is versatile and applicable to various aggregation networks. As shown in \cref{tab:ablation}, we apply it to 2D aggregation networks composed of 2D convolutions (BANet-2D) and 3D aggregation networks composed of 3D convolutions (BANet-3D), respectively. Compared to a single-branch aggregation network (Baseline), our bilateral aggregation (BA) can adaptively divide the image into high-frequency detail regions and low-frequency smooth regions, and then aggregate them accordingly. Without SSA, the attention map $\mathbf{A}$ is generated from the 1/4 scale feature $\mathbf{F}_{l,4}$. As a result, BA provides significant improvements with minimal computational cost. 

To more accurately identify high-frequency and low-frequency regions in the image, we propose a scale-aware spatial attention module that learns the differences in multi-scale image features to generate a spatial attention map for effectively separating high-frequency details and edges, and low-frequency smooth regions. \cref{tab:ablation} shows that scale-aware spatial attention can further boost performance.

We also present the ablation results on the test sets of KITTI 2012~\cite{kitti2012} and 2015~\cite{kitti2015}, as shown in \cref{tab:ablation_kitti}. Compared to the improvements on the synthetic Scene Flow test set, our bilateral aggregation achieves more significant gains on the challenging real-world test sets of KITTI 2012 and 2015. We highlight the improvements in the foreground regions (D1-fg) on KITTI test sets, as these regions usually contain more high-frequency details and edges. Specifically, for the D1-fg metric, our bilateral aggregation improves accuracy by 17\% for 2D convolution-based aggregation and 22\% for 3D convolution-based aggregation. 

Qualitative results are shown in \cref{fig:ba}. The attention map distinguishes high-frequency details and edges, and our bilateral aggregation preserves fine structures while ensuring accurate matching in textureless regions.

\begin{figure*}
\centering
\includegraphics[width=0.85\textwidth]{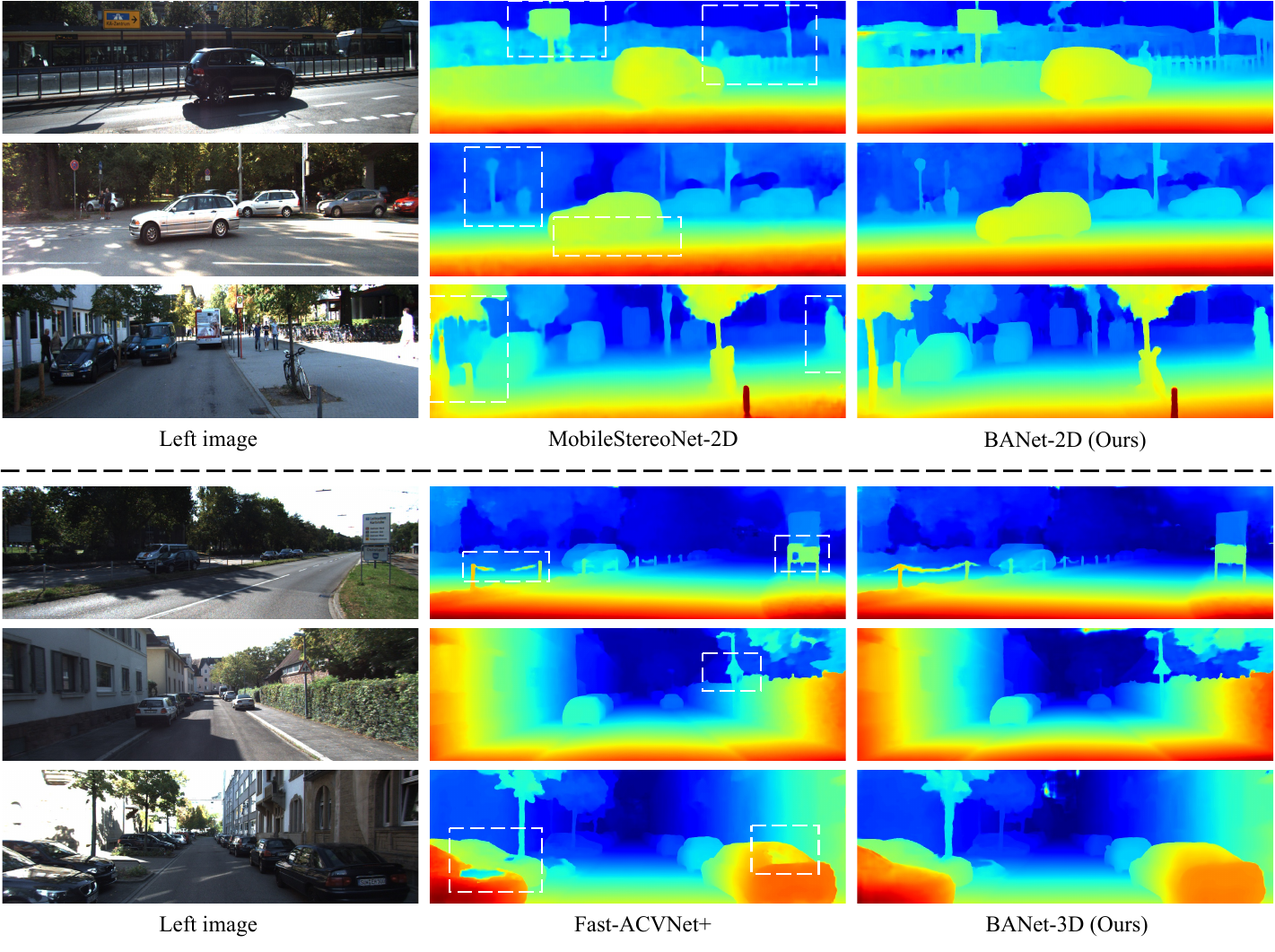} 
\caption{Qualitative comparisons on the test sets of KITTI 2012~\cite{kitti2012} and 2015~\cite{kitti2015}. 
By employing this divide-and-conquer approach, our bilateral aggregation produces clear edges and preserves intricate details, enabling accurate matching even in large textureless regions.}
\label{fig:kitti}
\end{figure*}

\begin{table}
  \centering
  \begin{tabular}{l|cc}
    \toprule
    Method & EPE (px) & Bad 3.0 (\%) \\ 
\midrule
PSMNet~\cite{psmnet} & 1.09 & 4.68 \\
BA+PSMNet & \textbf{0.77} & \textbf{3.28} \\
\midrule
GwcNet~\cite{gwcnet} & 0.76 & 3.30 \\
BA+GwcNet & \textbf{0.67} & \textbf{2.89}\\
\midrule
Fast-ACVNet+~\cite{fast-acv} & 0.59 & 2.70 \\
BA+Fast-ACVNet+ & \textbf{0.53} & \textbf{2.25} \\
\bottomrule
\end{tabular}
  \caption{
  Performance of Bilateral Aggregation (BA). Our BA can be seamlessly integrated into cost-volume aggregation methods, significantly enhancing their performance.}
  \label{tab:universality}
\end{table}

\subsection{Comparisons with State-of-the-Art Methods}

\noindent\textbf{Quantitative Comparisons.} \cref{tab:sceneflow} and \cref{tab:kitti_benchmark} present quantitative comparison results on the Scene Flow, KITTI 2012, and KITTI 2015 test sets. Our BANet-3D achieves the highest accuracy among the published real-time methods~\cite{bgnet,coex,hitnet,fast-acv,mobilestereonet} on high-end GPUs for almost all metrics. However, due to the use of 3D convolutions, BANet-3D is challenging to deploy on mobile devices. Without the use of 3D convolutions, our BANet-2D is more mobile-friendly and easy to deploy on mobile platforms. Although it is slightly inferior to BANet-3D, it surpasses all other lightweight methods. Specifically, on the KITTI 2015 test set, BANet-2D surpasses MobileStereoNet-2D~\cite{mobilestereonet} by 35.3\%, and BANet-3D outperforms FastACVNet+~\cite{fast-acv} by 11.9\% for the D1-all metric. 

Previous methods~\cite{hitnet,fast-acv,coex,bgnet,aanet} reported their runtime on their respective GPUs; however, the runtime can vary across different GPUs. To ensure a fair comparison of the computational complexity across methods, we uniformly measure MACs (Multiply-Accumulate Operations), which remain consistent regardless of GPU model. Our BANet-2D achieves the lowest MACs among all methods. 

\vspace{2mm}
\noindent\textbf{Qualitative Comparisons.} We compared the visual results of our methods with the mobile-friendly 2D convolution-based MobileStereoNet-2D~\cite{mobilestereonet} and the state-of-the-art 3D convolution-based Fast-ACVNet+~\cite{fast-acv}. As shown in \cref{fig:kitti}, a single-branch aggregation network often struggles to effectively handle both high-frequency edges and details, as well as large textureless regions, resulting in edge blurring, detail loss, and mismatches in textureless regions. In contrast, by employing this divide-and-conquer idea, our proposed bilateral aggregation produces clear edges and preserves intricate detail structures, resulting in accurate matching in large areas of textureless regions. In particular, our performance gains are more pronounced when integrated into simpler 2D convolution-based methods.

\vspace{2mm}
\noindent\textbf{Latency on Mobile Device.} We compare the latency with the latest mobile-friendly method, MobileStereoNet-2D~\cite{mobilestereonet}, on the Qualcomm Snapdragon 8 Gen 3, as shown in \cref{fig:teaser}. Benefiting from the proposed bilateral aggregation, our BANet-2D achieves sharp edges and accurate matching in textureless regions, while requiring only 45ms, which is less than one-third of MobileStereoNet-2D’s latency. Furthermore, we present a detailed breakdown of the latency: 16ms for feature extraction, 6.5ms for correlation volume construction, and 22.5ms for bilateral aggregation. 

\subsection{Universality and Superiority of BA}
To demonstrate the universality and superiority of the proposed bilateral aggregation, we integrate it into three representative methods, namely PSMNet~\cite{psmnet}, GwcNet~\cite{gwcnet}, and Fast-ACVNet+~\cite{fast-acv}, and compare the performance of the original methods with their counterparts enhanced by bilateral aggregation. The comparison results are presented in \cref{tab:universality}. Our bilateral aggregation significantly enhances those cost-volume aggregation-based methods, such as improving Fast-ACVNet+ by 10.2\% in the EPE metric.

%% file: sec/5_conclusion.tex
\section{Conclusion}
\label{sec:conclusion}

This paper presents a novel bilateral aggregation network for mobile stereo matching that achieves high-quality results using only 2D convolutions.
To effectively handle both detailed and smooth regions, we propose bilateral aggregation, which separates the full cost volume into detailed and smooth cost volumes, and then performs detailed and smooth aggregations accordingly. To more accurately distinguish between detailed and smooth regions, we propose a new scale-aware spatial attention module. Experimental results demonstrate that our method can run in real-time on mobile devices with high prediction accuracy, significantly outperforming existing methods.

An exciting future direction could involve extending our approach to other aggregation-based tasks, \eg multi-view stereo and optical flow estimation. Additionally, we believe that our mobile-friendly design could offer significant advantages for practical applications, such as drone navigation and intelligent photography.

\textbf{Acknowledgements.} This research is supported by the National Natural Science Foundation of China (623B2036, 62472184), National Key R\&D Program of China(2024YFE0217700), and the Fundamental Research Funds for the Central Universities.

%% file: sec/X_suppl.tex
\clearpage
\setcounter{page}{1}
\maketitlesupplementary

\begin{table}
  \centering
  \begin{tabular}{l|cccc}
    \toprule
    \multirow{2}{*}{Method} & \multicolumn{4}{c}{KITTI 2012 (Reflective)} \\
    & 3-noc & 3-all &4-noc &4-all \\
    \midrule
    BGNet+~\cite{bgnet} & 6.44 & 8.41 & 4.26 & 5.80 \\
    AANet+~\cite{aanet} & 7.22 & 9.10 & 5.25  & 6.66 \\
    CoEx~\cite{coex} & 6.83 & 8.63 & 4.61 & 6.00 \\
    Fast-ACVNet+~\cite{fast-acv} & 6.82 & 8.59 & 4.83 & 6.06 \\
    HITNet~\cite{hitnet} & 5.91 & 7.54 & 4.04 & 5.34 \\
    \midrule
    w/o BA (2D) & 8.81 & 10.61 & 6.17 & 7.63 \\
    BANet-2D & \underline{5.59} & \underline{7.27} & \underline{3.77} & \underline{5.08} \\
    \midrule
    w/o BA (3D) & 6.10 & 8.06 & 4.14 & 5.63 \\
    BANet-3D & \textbf{5.37} & \textbf{7.07} & \textbf{3.64} & \textbf{4.89} \\
    \bottomrule
  \end{tabular}
  \caption{Quantitative evaluation in reflective (ill-posed) regions of the KITTI 2012 test set.} 
  \label{tab:reflective}
\end{table}

\begin{figure}
\centering
{\includegraphics[width=1.0\linewidth]{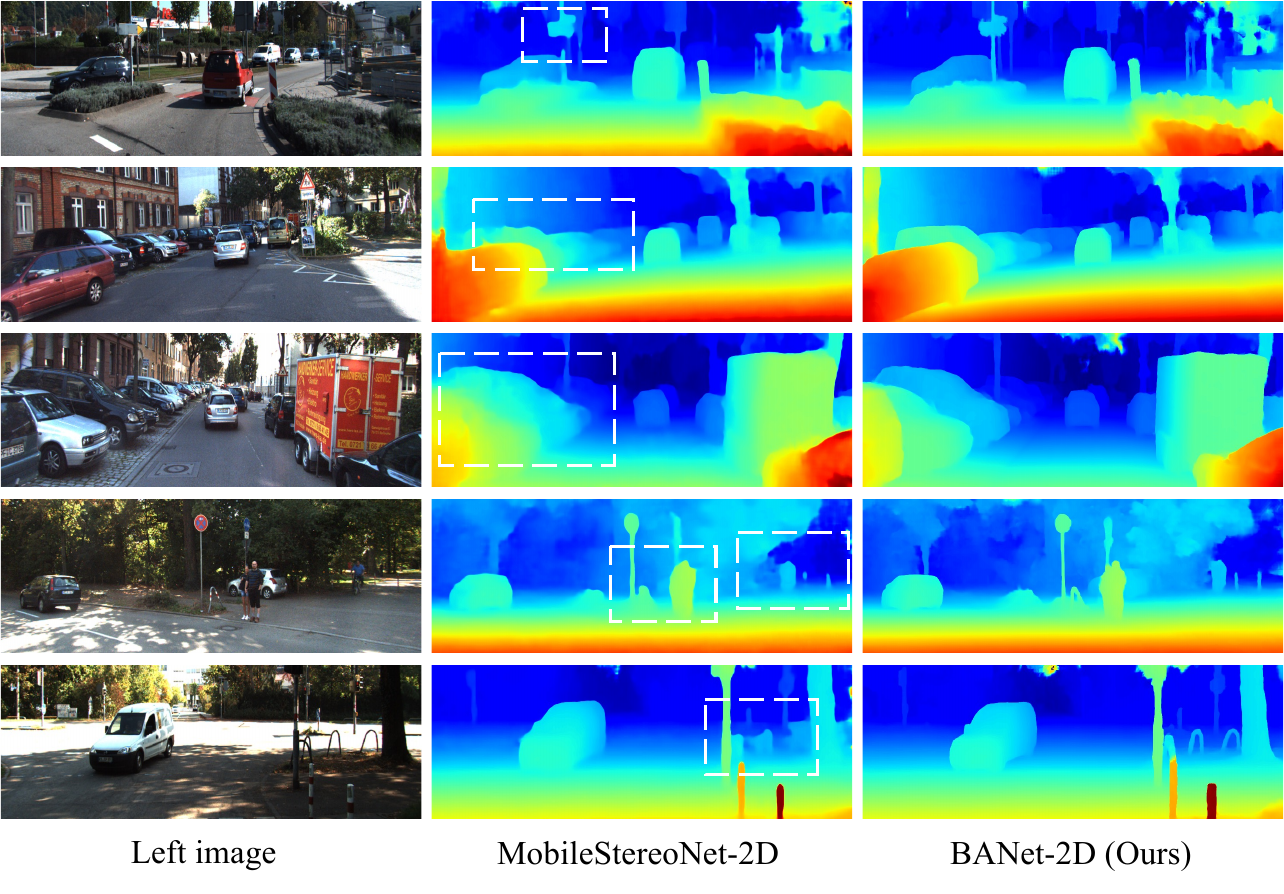}}
\caption{Qualitative comparisons with MobileStereoNet-2D~\cite{mobilestereonet} on the test set of KITTI 2015~\cite{kitti2015}. Significant improvements are highlighted by white dashed boxes. Our
bilateral aggregation produces clear edges and preserves intricate details. Zoom in for a clearer view.}
\label{fig:supp_v1}
\end{figure}

\begin{figure}
\centering
{\includegraphics[width=1.0\linewidth]{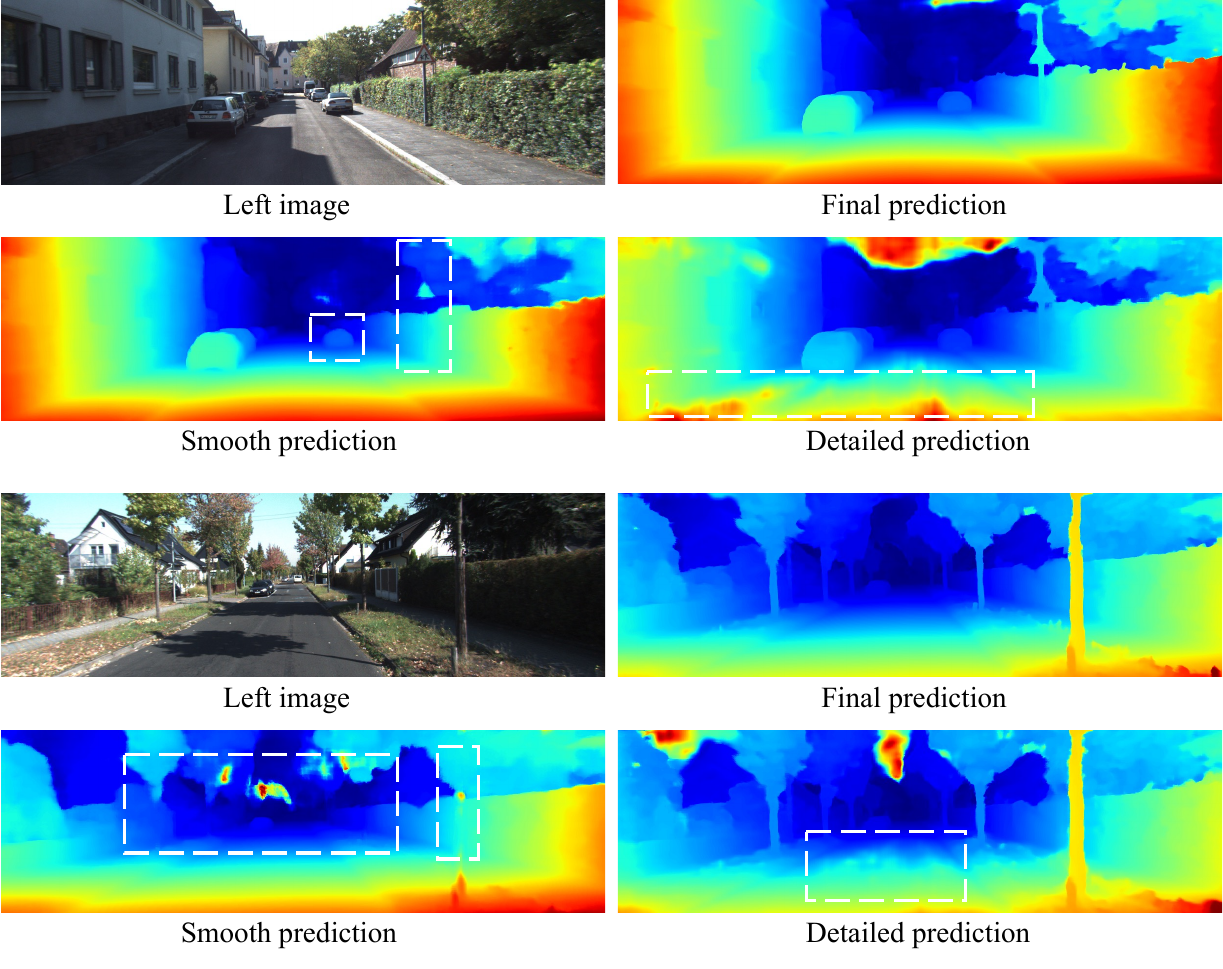}}
\caption{Visual results of the detailed aggregation branch, the smooth aggregation branch, and the final prediction.}
\label{fig:supp_v2}
\end{figure}

\section{More Experimental Results}

\subsection{Performance in Reflective Regions}
To verify the performance of our bilateral aggregation (BA) in smooth regions, such as reflective regions. 
We compared our approach with previously published real-time methods (on high-end GPUs) as well as our baseline method, which removes the bilateral aggregation (denoted as w/o BA). The comparison results are presented in \cref{tab:reflective}, where our BANet-3D achieves the best performance. Compared to other methods, our bilateral aggregation can adaptively separate the full cost volume into detailed and smooth cost volumes. This enables the smooth aggregation network to specialize in handling smooth regions, including reflective regions. As a result, our method outperforms previous approaches by a significant margin in reflective regions. 

Specifically, compared to the baseline (w/o BA), our bilateral aggregation (BA) achieves a significant improvement, such as a 36.5\% improvement for 2D convolution-based aggregation networks and a 12.0\% improvement for 3D convolution-based ones.

\subsection{More Qualitative Comparisons}
As shown in \cref{fig:supp_v1}, we provide more visual comparisons with MobileStereoNet-2D~\cite{mobilestereonet}. Both MobileStereoNet-2D and our BANet-2D rely solely on 2D convolutions, avoiding costly 3D convolutions or operations that are unfriendly to mobile devices. MobileStereoNet-2D struggles to simultaneously handle high-frequency edges and details, and low-frequency smooth regions, resulting in blurred edges and loss of fine details. In comparison, our method effectively addresses these challenges by the proposed bilateral aggregation. As a result, our approach produces clear edges and preserves intricate details.

\subsection{Detailed and Smooth Visual Results}
\cref{fig:supp_v2} shows visual results of the detailed aggregation branch, the smooth aggregation branch, and the final prediction. The smooth aggregation branch performs well in handling low-frequency smooth regions but struggles with high-frequency edges and details, as indicated by the white dashed boxes in \cref{fig:supp_v2}. In contrast, the detailed aggregation branch excels at handling high-frequency edges and details but underperforms in low-frequency smooth regions. Our final prediction combines the strengths of both branches, effectively addressing high-frequency edges and details while also handling low-frequency smooth regions.